\documentclass[final,5p,times]{elsarticle}

\journal{Systems \& Control Letters}

\usepackage{amssymb,amsmath,verbatim,bm,color,mathtools,multicol}
\usepackage{graphicx}
\usepackage{psfrag}
\usepackage{indentfirst}
\usepackage{hyperref} 
\usepackage{newtxtext}
\usepackage{algorithm}
\usepackage[noend]{algpseudocode}
\usepackage{siunitx}
\usepackage{tabularx}
\usepackage{multirow,array,booktabs}
\usepackage{caption}
\usepackage{subcaption}
\usepackage{wrapfig}
\usepackage{textcomp}
\usepackage{ulem}

\newcommand{\expect}[1]{\mathbb{E}_{#1}}

\newcommand{\kullback}{\mathbb{D}}
\newcommand{\kullbacks}[2]{\mathbb{D}\left[#1\left\|#2\right]\right.}
\newcommand{\reny}[2]{\mathbb{D}_\alpha\left[#1\left\|#2\right]\right.}
\newcommand{\matrixstyle}[1]{\mathrm{#1}}

\usepackage{graphicx}

\usepackage{tikz}
\usetikzlibrary{fit,positioning,automata,backgrounds}

\newtheorem{lemma}{Lemma}
\newtheorem{example}{Example}
\newtheorem{remark}{Remark}
\newtheorem{proposition}{Proposition}
\newproof{proof}{Proof}

\setcitestyle{square}

\begin{document}
	
	\begin{frontmatter}
		
		
		
		\title{{\color{black}Probabilistic Control and Majorisation of Optimal Control}}
		
		
		\author[label1,label2]{Tom Lefebvre}
		
		\address[label1]{Dynamic Design Lab, Department of Electromechanical, Systems and Metal Engineering, Ghent University, 9000 Ghent, Belgium.}
		\address[label2]{FlandersMake@UGent -- corelab MIRO, 9052 Ghent, Belgium.}
		
		\begin{abstract} 
			{\color{black}Probabilistic control design is founded on the principle that a rational agent attempts to match modelled with an arbitrary desired closed-loop system trajectory density. }{\color{black} The framework was originally proposed as a tractable alternative to traditional optimal control design, parametrizing desired behaviour through fictitious transition and policy densities and using the information projection as a proximity measure. In this work we introduce an alternative parametrization of desired closed-loop behaviour and explore alternative proximity measures between densities.} {\color{black}It is then illustrated} how the associated probabilistic control problems solve into uncertain or probabilistic policies. Our main result is to show that the probabilistic control {\color{black}objectives} majorize conventional, stochastic and risk sensitive, optimal control objectives. This observation allows us to identify two probabilistic fixed point iterations that converge to the deterministic optimal control policies {\color{black} establishing an explicit connection between either formulations. Further we demonstrate that the risk sensitive optimal control formulation is also technically equivalent to a Maximum Likelihood estimation problem on a probabilistic graph model where the notion of costs is directly encoded into the model. The associated treatment of the estimation problem is then shown to coincide with the moment projected probabilistic control formulation. That way optimal decision making can be reformulated as an iterative inference problem.} Based on these insights we discuss directions for algorithmic development.
		\end{abstract}
		
	\end{frontmatter}
	
	\section{Introduction}
	\vspace*{-5pt}
	
	{\color{black}Optimal control is a mathematical formalization tailored to intelligent decision making \cite{aastrom2012introduction,bertsekas2012dynamic}.} {\color{black} A rational agent makes policy such that, in each situation, a decision is made that is expected to accumulate a minimal cost over a finite (or infinite) time horizon. {\color{black} To that end, the agent represents the dynamic system with a probabilistic model and introduces an external notion of cost that quantifies the value of every closed-loop behaviour accounted for by the model. The agent chooses a policy that is associated to a minimal cost on average.} By (meticulous) design of the cost function it is possible to encode complex behavioural {\color{black}tendencies} into the policy {\color{black}that will manifest in the closed-loop system} \cite{mordatch2012trajopt,posa2014trajopt,mastalli2020crocoddyl}. Such optimal policies are an elegant theoretical concept but are typically hard to come by in practice. The optimal policy is characterised by a recursive dynamic programming equation. As a result, explicit expressions for the optimal policy exist only for a handful of problem statements. 
		
		{\color{black} Probabilistic control was proposed by K\'{a}rn\'{y} \cite{karny1996towards,karny2006fully,karny2012axiomatisation,karny2020axiomatisation} as an alternative formulation for control design resulting in a more tractable framework. In optimal control the minimization of the expected cost can be interpreted as an attempt to influence selected characteristics of the closed-loop density. Deriving from this observation, probabilistic control presents an alternative control design formulation. A rational agent is said to act so that its modelled closed-loop density is forced to to be as close as possible to some desired density. The proximity between densities was measured by the relative entropy between the modelled and desired behaviour. Specific to this formulation is that the explicit form of the probabilistic controller can be found where the probabilistic policy is still governed by a recursion however the minimization operator is replaced by an expectation.}

	}

	{	
		{	\color{black}A closely related research program was carried out by the robotics and Reinforcement Learning community, attempting to wield inference principles in control. Several authors have aimed to encode the notion of reward directly into the probabilistic model that is used to represent the system \cite{attias2003planning,toussaint2006probabilistic,toussaint2009robot,hoffmann2017linear}. An auxiliary set of exogenous binary \textit{optimality variables} is introduced whose values are assumed to be known and true, indicating that an optimal decision has been made. The agent infers the most likely action assuming future optimality has been achieved. That way, optimal decision making could be formalized as an inference problem }{\color{black} and if successful this would allow to bring to bear a wide range of inference techniques to address the optimization itself. Other authors have pursued the introduction of entropic principles and entropic regularization terms into the optimal control objective \cite{ziebart2010modelingb,kappen2012optimal,levine2013variational,rawlik2013stochastic}. The body of work accumulated into a paradigm that is now often referred to as \textit{Control as Inference} (CaI) \cite{levine2018reinforcement,abdolmaleki2018maximum}. 
		}
		
		All of these attempts culminate into roughly the same underlying framework and optimization objective. What remains unclear however is how these problems relate to the theory of optimal control{ \color{black}and what advantages they have to offer over the latter.} In this context this article has three main aims
		\begin{enumerate}
			\item Our first aim is to give a concise introduction to the body of work underpinning probabilistic control and control as inference. Therefore we extend on the {density} matching formulation from probabilistic control. {\color{black}In this setting we propose an alternative parametrization of the desired closed-loop density and explore the use of other proximity measures between densities. These steps proof to be crucial to establish a comprehensive connection with optimal control theory and probabilistic control, hence CaI.}
			\item {\color{black}Second we establish an explicit connection between two probabilistic control and optimal control formulations.} In particular we show that the {\color{black}information projected and moment projected} probabilistic control objectives majorize stochastic and risk sensitive optimal control objectives {\color{black}respectively}. Thereby we identify {\color{black}the probabilistic formulations} to single out a step in a fixed point iteration {\color{black} that maintains a series of probabilistic policies and whose stationary point coincides with the deterministic policy.} 
			{\color{black}
				\item We further show that the moment projected probabilistic control policy, proposed in this contribution and that we illustrate is associated to risk sensitive optimal control, is also technically equivalent to the calculation of the probability of the control conditioned on the state and future \textit{optimality variables}. This observation renders it equivalent to message passing on a probabilistic graph model. To establish this technical equivalence it is discussed how the risk sensitive optimal control problem is technically equivalent to a Maximum Likelihood estimation corresponding to the extended probabilistic (graph) model.}
		\end{enumerate}
		
		In conclusion we give a brief outlook and sketch blueprints for further algorithmic development building on the technical results given and exploiting the rich technical machinery tailored to efficient approximate probabilistic inference.
	}
	
	\section{Background}
	
	In this section we give a brief primer on some key models and concepts required to develop our main {\color{black}results}. First remark that our treatment will be restricted to discrete time. Further, our presentation is limited to continuous spaces. Generalization to discrete spaces is trivial and the results valid in either setting.
	
	{\color{black}
		\subsection{Notation}
		Notations, $\underline{x}_t = \{x_0,\dots,x_t\}$, and, $\overline{x}_t = \{x_t,\dots,x_T\}$, refer to a leading or trailing part of a sequence, $t$ refers to the final or initial time instance of the corresponding subsequence. Subindex $t$ represents discrete time, $T\in\mathbb{N}$ is some natural number associated to the total length of sequences. We silently assume that a complete sequence starts at time $t=0$ and ends at time $t=T$. Throughout we refer to the set of all feasible probability density functions with $\mathcal{P}$. We will rely on the context to imply the arguments and properties of the corresponding function class.
	}
	
	{\color{black}\subsection{Agent and descriptive model}
		
		We refer to the entity tasked with the decision process, ergo with making policy, as the agent. It is assumed that the agent \textit{makes sense} of the system by means of a veridical probabilistic descriptive model. Specifically the agents models the controlled process} as a Controlled Markov Chains (CMCs). 
	
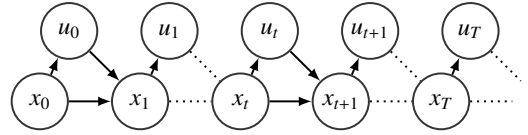
\begin{figure}[t] 
		\centering 
		\resizebox{.85\columnwidth}{!}{
			\begin{tikzpicture}
				\tikzstyle{main}=[circle, minimum size = 8mm, thick, draw =black!80, node distance = 6mm]
				\tikzstyle{connect} = [-latex,thick]
				\tikzstyle{floating} = [thick,dotted]
				\tikzstyle{measure} = [-latex,thick,bend right=45]
				
				\node[main] (X0) {$x_0$};
				\node[main] (X1) [right=of X0] {$x_1$};
				\node[main] (Xt) [right=of X1] {$x_t$};
				\node[main] (Xtt) [right=of Xt,label=center:$x_{t+1}$] {};
				\node[main] (XT) [right=of Xtt] {$x_{T}$};
				\node[main,draw=white] (XTT) [right=of XT] {};
				
				\node[main] (U0) [above left=of X1] {$u_0$};
				\node[main] (U1) [right=of U0] {$u_1$};
				\node[main] (Ut) [right=of U1] {$u_t$};
				\node[main] (Utt) [right=of Ut,label=center:$u_{t+1}$] {};
				\node[main] (UT) [right=of Utt] {$u_{T}$};
				
				\path (X0) edge [connect] (X1)
				(X1) edge [floating] (Xt)
				(Xt) edge [connect] (Xtt)
				(Xtt) edge [floating] (XT)
				(XT) edge [floating] (XTT);
				
				\path (X0) edge [connect] (U0);
				\path (X1) edge [connect] (U1);
				\path (Xt) edge [connect] (Ut);
				\path (Xtt) edge [connect] (Utt);
				\path (XT) edge [connect] (UT);
				
				\path (U0) edge [connect] (X1);
				\path (U1) edge [floating] (Xt);
				\path (Ut) edge [connect] (Xtt);
				\path (Utt) edge [floating] (XT);
				\path (UT) edge [floating] (XTT);	
			\end{tikzpicture}
		}
		\caption{Graphical representation of a Controlled Markov Chain.} 
		\label{fig:CMC} 
	\end{figure}

	The conditional dependencies that govern a CMC are conveniently illustrated by the probabilistic graph model in Fig. \ref{fig:CMC}. {\color{black}The process  $x_0$, ..., $x_T$, represents the system's states and assumes values in the set $\mathcal{X}$. Every time $t$, the agent can act on the system by deciding and applying a control, $u_t$.} The control variables or actions, $u_0$, \dots, $u_T$, assume values in the set, $\mathcal{U}$. {\color{black}We further define the tuple, $\xi_t=(x_t,u_t)$, to alleviate notation.}
	\newpage
	
	The agent's model, $\mathcal{M}$, is characterised by an initial model, {\color{black}$p(x_0) = \iota(x_0)$}, a transition model, $p(x_{t+1}|x_t,u_t) = \tau_t(\xi_t,x_{t+1})$ and \textit{a policy model}, $\rho_t(\xi_t) = p(u_t|x_t)$. For now the choice of action, $u_t$, as a function of the present state, $x_t$, is not governed by the intention of rendering some utility function optimal. Instead, the choice is subject to a sequence of uncertain decision functions, i.e. $\underline{\rho}_T$. It is assumed we can define such mapping arbitrarily. 
	
	Following the probabilistic graph model in Fig. \ref{fig:CMC}, the joint model can be decomposed in terms of the local models. Specifically, the uncertainty about a \textit{state-action} trajectory, $\underline{\xi}_T$, is determined by the product of all transition and policy models.
	\begin{equation} 
		p(\underline{\xi}_T;\underline{\rho}_{T-1}) = {\color{black}\iota(x_0)} \prod\nolimits_{t=0}^{T-1} \tau_t(\xi_t,x_{t+1}){\color{black}\rho_t(\xi_t) }
	\end{equation} 
	
	Given the crucial role that the policy models will play throughout, we say, and emphasise notationally, that the trajectory density is \textit{parametrised} by the policy sequence, $\underline{\rho}_{T-1}$. 
	
	Further note that the parametrized trajectory density describes the entire set of feasible behaviours of the system, $\mathcal{F}$.
	\begin{equation}
		\mathcal{F}_\rho = \left\{p(\underline{\xi}_T;\underline{\rho}_{T-1}) \left|  \underline{\rho}_{T-1} \in \mathcal{P}\right\} \right.
	\end{equation}
	
	{\color{black}
		{\color{black}\subsection{Optimal control and common decision criteria}\label{sec:optimal-control-problems}}

		{\color{black}
			Control design for stochastic systems is traditionally based on optimization of the expected value of a suitably chosen cost. There exist two measures of cost that produce a rich mathematical theory and that have been studied extensively as a result. 
			
			The first is given by the additive cost, $A$. Note that we extend the notation systems introduced before to imply the terms that are incorporated in a sum. For example, $\underline{c}_t$, refers to the leading terms, whilst $\overline{c}_t$ refers to the trailing terms.} 
		\begin{equation}
			A = \underline{c}_T(\underline{\xi}_T) = \sum\nolimits_{t=0}^{T-1} c_t(\xi_t) + c_T(x_T)
		\end{equation}
		
		The standard Stochastic Optimal Control (SOC) objective, $a$, can then be defined concisely, {\color{black}adopting the descriptive model from the previous section}. The cost in (\ref{eq:SOC}) formalises many methodologies tailored to control, such as trajectory generation, model based predictive control, reinforcement learning, etc. 
		\begin{equation}
			\label{eq:SOC}
			a[\underline{\pi}_{T-1}] =  \expect{p(\underline{\xi}_T;\underline{\pi}_{T-1})}\left[\underline{c}_T(\underline{\xi}_T) \right]
		\end{equation}
		
		{\color{black}Alternative to the addictive cost, $A$, one can consider the multiplicative cost, $M$, which recycles the additive cost, $A$, and makes use of the properties of the exponential to render the associated cost structure multiplicative. In terms of utility theory, this means that high costs are more or less (depending on $\sigma$ ) advantageous to the agent then $A$ leads to suspect.
			\begin{equation}
				M = \exp(-\sigma A) = \exp(-\sigma \underline{c}_T(\underline{\xi}_T) )
			\end{equation}
			
			The multiplicative cost is associated with the Risk-Sensitive Optimal Control (RSOC) objective \cite{whittle1996optimal,whittle1981risk,whittle2002risk}.} The solution is referred to as \textit{risk averse} for $\sigma < 0$ and \textit{seeking} for $\sigma > 0$. In the limit $\sigma\rightarrow0$, the RSOC collapses onto the SOC problem. We further absorb $\sigma$ into $\underline{c}_T$ which can be achieved through appropriate scaling. {\color{black}In this work we focus on the risk seeking version for reasons that will be explained later.}
		\begin{equation}
			\label{eq:RSOC}
			m[\underline{\pi}_{T-1}] = - \frac{1}{\sigma}\log \expect{p(\underline{\xi}_T;\underline{\pi}_{T-1})}\left[\exp(-\sigma \underline{c}_T(\underline{\xi}_T) )\right]
		\end{equation}
		
			Remark that we have formulated optimal control relying on the trajectory density, $p(\underline{\xi}_T;\underline{\pi}_{T-1})$, which is parametrised by some arbitrary sequence of policies, $\underline{\pi}_{T-1}$, living in the set of densities conditioned on $\underline{x}_T$. This may appear to suggest -- incorrectly -- that the solution of either of these problems is governed by an uncertain sequence of decision functions. We emphasize that neither objective, $a$ nor $m$, is rendered extreme by an uncertain policy, rather the solution is given by a sequence of optimal deterministic policy functions, $\pi_t^*:\mathcal{X}\mapsto\mathcal{U}$. 
			Regardless both problems can be treated {probabilistically}, with the optimal solution {collapsing} onto a Dirac delta function. Since the set of all deterministic decision functions is a subset of the uncertain set, the present formulation only extends the optimisation space.}
		
	
	\vspace*{5pt}
	\subsection{Information-Theoretic projections}\label{sec:information-theoretic-projections}
	
	{\color{black}The Bayesian interpretation of probability theory} is that the state of knowledge, a.k.a. the belief, of the reasoning entity {\color{black}is associated with} the probability over all possible outcomes. 
	Inference principles govern how to update a prior belief into a posterior belief when new information becomes available and the state of knowledge changes. Bayesian inference can be used to process information that is represented by the outcome of experiments, i.e. empirical evidence. The Information- and Moment-projection, respectively abbreviated to the I- and M-projection, are information-theoretic concepts that can be used to process information represented by constraints that affect the belief space. We could refer to such evidence as {structural}. 
	
	Both concepts are based on the relative entropy 
	\begin{equation}
		\kullbacks{\pi}{\rho} = \expect{\pi}\left[\log \tfrac{\pi}{\rho}\right]
	\end{equation}
	between densities $\pi\in\mathcal{P}$ and $\rho\in\mathcal{P}$. 
	
	The relative entropy is a measure of the inefficiency of assuming that the density is $\rho$ when the true density is $\pi$ \cite{cover2006elements}. Another interpretation defines it, tautologically, as a quantitative measure of the magnitude of the update from a prior, $\rho$, to a posterior, $\pi$ \cite{pml2Book}. The latter interpretation leads directly to the principle of minimum relative entropy {\color{black}and related entropic inference}, advocated by Jaynes amongst others \cite{jaynes1982rationale,caticha2011entropic}, which states that the relative entropy is to be minimized if we want to encode some form of new information affecting our belief (usually an expectation of the form $\expect{\pi}[f] = \mu$) into the prior $\rho$. 
	
	\begin{enumerate}
		\item\textbf{I-projection} The I-projection is equivalent with $\pi^\bullet$ the I-projection of $\rho$ onto $\mathcal{P}^*$.
		\begin{equation*}
			\pi^\bullet = \arg\min\nolimits_{\pi\in\mathcal{P}^*} \mathbb{D}[\pi\parallel\rho] 
		\end{equation*}
		\item\textbf{M-projection} The M-projection is the reciprocal of the I-projection with $\pi^\star$ the M-projection of $\rho$ onto $\mathcal{P}^*$.
		\begin{equation*}
			\pi^\star=\arg\min\nolimits_{\pi\in\mathcal{P}^*} \mathbb{D}[\rho\parallel\pi] 
		\end{equation*}
	\end{enumerate}
	Here the set $\mathcal{P}^*\subset\mathcal{P}$ represents the constrained belief space, e.g. any belief that satisfies $\expect{\pi}[f] = \mu$, i.e. $\mathcal{P}^* = \{p\in\mathcal{P}|\expect{p}[f] = \mu\}$. 
	
\begin{figure}[t]
		\centering
		\begin{subfigure}[b]{0.3\columnwidth}
			\centering
			\includegraphics[width=\columnwidth]{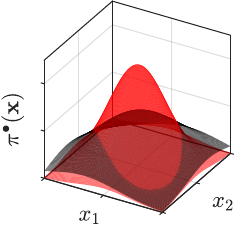}
			\caption{}
		\end{subfigure}
		\begin{subfigure}[b]{0.3\columnwidth}
			\centering
			\includegraphics[width=\columnwidth]{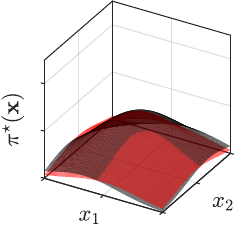}
			\caption{}
		\end{subfigure}
		\caption{Illustration of mode-seeking (a) and mode-covering behaviour (b) of the I- and M-projection respectively. The black surface represents the projected density $\rho$. The red surfaces represent the projection,  $\pi$, on the family of homogeneous bivariate normal densities. {\color{black} Remark how $\pi^\bullet$ concentrates on the smallest eigenvalue of the covariance of $\rho$ whereas $\pi^\star$ concentrates on the highest eigenvalue.} Figures are adopted from \cite{pml2Book}. }
		\label{fig:IM}
	\end{figure}

	The relative entropy is a divergence and not a distance and thus asymmetric in its arguments. Therefore the I-projection and the M-projection do not yield the same projection \cite{bishop2006pattern,pml1Book}. They are either \textit{mode seeking} or \textit{covering} for $\pi$. As a result the I-projection will underestimate the support of $\rho$ and vice versa. This behaviour is illustrated in Fig. \ref{fig:IM} where a heterogeneous Gaussian is projected on the set of homogeneous Gaussians.
	
	\section{Probabilistic Control}\label{sec:probabilistic-control}
	
	{\color{black}In this section we review and extend the notion of probabilistic control theory as an alternative framework to optimal control. To that end we will first formulate the principle in natural language and then construct a mathematical framework that supports the principle. By reconstructing the framework from the underlying principle it is easy to see alternative manifestations of it that thus far have not been considered.}

	Formulated in terms of natural language, probabilistic control aims to find feedback control strategies, $\underline{\pi}_T$, so that the joint density of the closed-loop variables is forced to be as closely as possible to some desired trajectory density, $p^*(\underline{\xi}_T)$ \cite{karny1996towards,karny2006fully}. This idea can arguably be described as a density matching approach.
	
	This natural description introduces two questions
	\begin{enumerate}
		\item How does one define a purposeful desired density?
		{\color{black}		\item How does one measure the proximity of densities?}
	\end{enumerate}
	
	\subsection{Desired densities}
	
	The desired density is hypothetical and can be chosen arbitrarily {\color{black}as long as the result is a density with argument, $\underline{\xi}_T$.}  A productive {\color{black}parametrization} is given below. Here the desired density is proportional to the joint closed-loop variable model for some (arbitrary) prior feedback policy, $\underline{\rho}_T$, multiplied with a term that expresses the desirability of certain trajectories over others. 
	
	{\color{black}The desirability of a trajectory is encoded by the exponential of the negative additive cost, $A$. Likewise we multiply with the multiplicative cost, $M$. Normalization over $\underline{\xi}_T$ yields a proper density.}
	\begin{equation}
		\label{eq:desirable}
		p^*(\underline{\xi}_T;\underline{\rho}_{T-1}) \propto p(\underline{\xi}_T;\underline{\rho}_{T-1}) \exp(-\underline{c}_T(\underline{\xi}_T))
	\end{equation}
	
	{\color{black}Probabilistically}, the additive cost is transformed into of likelihood which attributes high probability to low cost trajectories and low probability to high cost trajectories. {\color{black}Through this} particular choice of likelihood, the {\color{black} formulation shall resume close analogies with optimal control though it is not less general than K\'{a}rn\'{y}'s orginal formulation.} 
	{\color{black}Here} the desired density takes the form of a hypothetical trajectory density consisting of desired transition and control densities
	\begin{equation}
		p^*(\underline{\xi}_T) = p(\underline{\xi}_T;\underline{\tau}_{T-1}^*,\underline{\rho}_{T-1}^*)
	\end{equation}
	
	Some remarks can be made that motivate the former definition is more productive than the latter
	
	\begin{remark}
		The latter density has the disadvantage that it may describe densities that are not achievable for the system because here we enforce a transition probability, $\tau_t^*$. To speak in behavioural terms, $p^*(\underline{\xi}_T)$ might be out of the set of feasible trajectory densities. The other goal density in (\ref{eq:desirable}) is in that sense favourable, because the presence of the true transition density makes sure that if certain behaviours are unachievable, they are unachievable for the desired density by default.
	\end{remark}
	
	\begin{remark}
		Note that it is possible to define an additive cost, $A$, so that both desired density models are equivalent
		\begin{equation}
			\underline{c}_T(\underline{\xi}_T) = -\sum\nolimits_{t=0}^{T-1} \log \frac{\tau_t^*(\xi_t)}{\tau_t(\xi_t)}
		\end{equation}
		This cost model is ill defined if the $\log$ of the ratio of the transition densities  is. Addressing this issue naturally leads to a description that is equivalent to that in (\ref{eq:desirable}).
	\end{remark}

	\begin{figure}[t]
		\centering
		\includegraphics[width=.9\columnwidth]{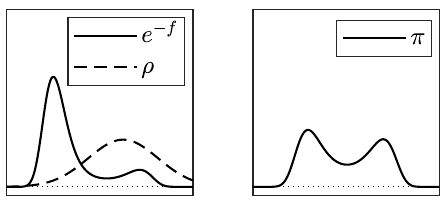}
		\caption{Illustration of (\ref{eq:desirable}) for $T=0$.}
		\label{fig:desirable}
	\end{figure}
	
	The concept of the desired density (\ref{eq:desirable}) is visualized in Fig. \ref{fig:desirable} for $T=0$, $x_T = u_{T-1}$ and $c_T(x_T) = f(x_T)$.
	\vfill
	
	\subsection{Defining proximity}
	
	To address the second question we are to define a suitable measure of proximity. {\color{black}To that end} we can rely on the information-theoretic projections introduced in section \ref{sec:information-theoretic-projections}. 
	
	\subsubsection{I-projection}
	The I-projection of $p^*$ onto $\mathcal{F}_\pi$ is formally defined as
	\begin{equation}
		\label{eq:Iproj}
		\underline{\pi}_{T-1}^\bullet = \arg\min_{\underline{\pi}_{T-1}} \kullbacks{p(\underline{\xi}_T;\underline{\pi}_{T-1})}{p^*(\underline{\xi}_T;\underline{\rho}_{T-1})}
	\end{equation}
	
	This problem definition was originally considered by K\'{a}rn\'{y} in \cite{karny1996towards}, and, rediscovered independently by other authors since, \cite{ziebart2010modelingb,kappen2012optimal,levine2013variational,rawlik2013stochastic}. {\color{black}In particular, problem (\ref{eq:Iproj}) is equivalent to the relative entropy optimal control objective where the additive cost measure is extended with the expected relative entropy between the prior policies, $\underline{\rho}_T$, and the optimal policy, $\underline{\pi}_T$.}
	
	The solution is governed by a recursive process.
	\begin{lemma} 
		\label{prop:Iproj}
		Consider the I-projection in (\ref{eq:Iproj}) and let $p^*$ be defined as in (\ref{eq:desirable}). Then the optimal control is given by
		\begin{equation*}
			\pi_t^\bullet(\xi_t) = {\rho}_{t}(\xi_t) \frac{\exp(-{Q}^\bullet_t(\xi_t))}{\exp(-{V}^\bullet_t({x}_t))}  
		\end{equation*}
		Starting with $V^\bullet_T = c_T$, the functions $V^\bullet_t$ and $Q_t^\bullet$ are generated recursively in a backward manner as follows
		\begin{equation*} 
			\begin{aligned} 
				V^\bullet_{t}({x}_t) &= -\log \expect{\rho_t(u_t|x_t)}\left[\exp(-{Q}^\bullet_t(\xi_t))\right] \\
				Q^\bullet_{t}(\xi_t) &= c_t({\xi}_t)+ \expect{p({x}_{t+1}|{\xi}_t)}\left[V^\bullet_{t+1}({x}_{t+1})\right]
			\end{aligned} 
		\end{equation*}
	\end{lemma}
	For the proof see \ref{sec:proof-of-proposition-refpropiproj}.
	
	\subsubsection{M-projection} The M-projection of $p^*$ onto $\mathcal{F}_\pi$, which is the information-theoretic reciprocal of (\ref{eq:Iproj}), is formally defined as
	\begin{equation}
		\label{eq:Mproj}
		\underline{\pi}_{T-1}^\star = \arg\min_{\underline{\pi}_{T-1}} 	\kullbacks{p^*(\underline{\xi}_T;\underline{\rho}_{T-1})}{p(\underline{\xi}_T;\underline{\pi}_{T-1})}
	\end{equation}
	
	As far as we are aware of this problem has not been treated by any other authors. Again the solution is governed by a recursion. 
	
	\begin{proposition} 
		\label{prop:Mproj}
		Consider the M-projection in (\ref{eq:Mproj}) and let $p^*$ be defined as in (\ref{eq:desirable}). Then the optimal control is given by
		\begin{equation*}
			\pi_t^\star(\xi_t) = {\rho}_{t}(\xi_t) \frac{\exp(-{Q}^\star_t(\xi_t))}{\exp(-{V}^\star_t({x}_t))} 
		\end{equation*}
		Starting with $V^\star_T = c_T$, the functions $V^\star_t$ and $Q_t^\star$ are generated recursively in a backward manner as follows
		\begin{equation*}
			\begin{aligned} 
				V^\star_{t}({x}_t) &= -\log \expect{\rho_t(u_t|x_t)}\left[\exp(-{Q}^\star_t(\xi_t))\right] \\
				Q^\star_{t}(\xi_t) &= c_t({\xi}_t)-\log \expect{p({x}_{t+1}|{\xi}_t)}\left[\exp(-V^\star_{t+1}({x}_{t+1}))\right] 
			\end{aligned} 
		\end{equation*} 
	\end{proposition}
	For the proof see \ref{sec:proof-of-proposition-refpropmproj}.
	
	\subsubsection{Generalisation through Rény divergences}\label{sec:generalisation-through-reny-divergences}
	A generalisation of the results above may be attempted through the study of the family of Rény divergences. The family of Rény divergences is defined (informally) as
	\begin{equation}
		\reny{p}{q} = \frac{1}{\alpha(\alpha-1)} \log \int p(x)^\alpha q(x)^{1-\alpha} \text{d}x
	\end{equation}
	
	The family is found to generalize the I- and M-projection
	\begin{equation}
		\begin{aligned}
			\lim_{\alpha\rightarrow 0} \reny{q}{p} &= \kullbacks{p}{q}\\ 
			\lim_{\alpha\rightarrow 1} \reny{q}{p} &= \kullbacks{q}{p}
		\end{aligned}
	\end{equation}
	
	Using the Rény divergence family, $\kullback_\alpha$, the proximity measures from the previous paragraphs can be generalised 
	\begin{equation}
		\label{eq:Rproj}
		\underline{\pi}_{T-1}^* = \arg\min_{\underline{\pi}_{T-1}} \reny{p^*(\underline{\xi}_T;\underline{\rho}_{T-1})}{p(\underline{\xi}_T;\underline{\pi}_{T-1})}
	\end{equation}
	
	\begin{proposition} 
		\label{prop:Rproj}
		Consider the Rény projection in (\ref{eq:Rproj}) and let $p^*$ be defined as in (\ref{eq:desirable}). Then the optimal control is given by
		\begin{equation*}
			\pi_t^*(\xi_t) = {\rho}_{t}(\xi_t) \frac{\exp(-{Q}^*_t(\xi_t))}{\exp(-{V}^*_t({x}_t))}
		\end{equation*}
		Starting with $V^*_T = c_T$, the functions $V^*_t$ and $Q_t^*$ are generated recursively in a backward manner as follows
		\begin{equation*} 
			\begin{aligned} 
				V^*_{t}({x}_t) &= -\log \expect{\rho_t(u_t|x_t)}\left[\exp(-{Q}^*_t(\xi_t))\right] \\
				Q^*_{t}(\xi_t) &= c_t({\xi}_t)-\tfrac{1}{\alpha}\log \expect{p({x}_{t+1}|{\xi}_t)}\left[\exp(-\alpha V^*_{t+1}({x}_{t+1}))\right] 
			\end{aligned} 
		\end{equation*} 
	\end{proposition}
	For the proof see \ref{sec:proof-of-proposition-refproprproj}. 
	
	Indeed these formulas are found to generalize the other schemes, noting that
	\begin{equation}
		\begin{aligned}
			\lim_{\alpha\rightarrow0} Q^*_t = Q^\bullet_{t} \\
			\lim_{\alpha\rightarrow1} Q^*_t = Q^\star_{t}
		\end{aligned}
	\end{equation}
	
	{\color{black}Lemma \ref{prop:Iproj} and propositions} \ref{prop:Mproj} and \ref{prop:Rproj} demonstrate that the probabilistic control problems (\ref{eq:Iproj}), (\ref{eq:Mproj}) and (\ref{eq:Rproj}) can be evaluated explicitly. {\color{black}This was, amongst other things, the original motivation to formulate and study probabilistic control problems. Indeed, although the calculation procedure is still recursive, it does not longer involve a minimization, therewith improving the tractability of the associated computation. On the other hand, it is less straightforward to intuitively grasp what effect open design choices, related to $\underline{\rho}_T$ and $\underline{c}_T$, have on the behavioural tendencies induced by the probabilistic control policies. Finally, the notion of a probabilistic control policy is also unsatisfactory from a decision making perspective, since the agent is still indecisive on what control to apply to the system. 
		
		To address these open issues, next we will, based on the proposed extension of proximity measure and our definition of desired densities, illustrate how probabilistic control relates to other control formulations, in particular optimal control.} 
	
	\section{Majorizing-Minimizing of Optimal Control problems}\label{sec:majorizing-minimizing-optimal-control}
	
	{\color{black} Definition of the desired density, $p^*$, by means of the likelihood transformation of the additive cost, $A$, is a clear indication that probabilistic and optimal control are closely related. The interpretation and choice of the prior policy, $\underline{\rho}_{T-1}$, rather than that it is arbitrary in some sense, is less straightforward.} Further, given the structure of the probabilistic control policies, an evident question is to ask what happens if we were to iterate the {solutions}? As it happens, the answer to this question proofs key to understand the relation between both formulations. 
	
	To establish the connection it is required {\color{black}to present a brief primer} on the Majorizing-Minimizing (MM) principle.
	
	
	\subsection{Majorizing-Minimizing}
	
	The MM principle aims to convert hard optimization problems into a sequence of easier optimization problems \cite{lange2016mm}. 
	\begin{equation}
		\min_\theta f(\theta)
	\end{equation}
	
	When the goal is to minimize the objective, as in the example above, the MM principle requires to majorize the objective function, $f(\theta)$, with a surrogate, $g(\theta,\theta')$, anchored at the current iterate, $\theta'$. Majorization of an objective imposes two requirements on the surrogate: (1) the tangency condition, and, (2) the domination condition, with $a>0$ and $b$ independent of $\theta$
	\begin{equation}
		\label{eq:MM}
		\begin{aligned}
			f(\theta) &\leq a \cdot g(\theta,\theta') + b \\
			f(\theta') &= a \cdot g(\theta',\theta') + b 
		\end{aligned}
	\end{equation}
	
	The surrogate is then used as a proxy for the true objective to obtain a new iterate through the following fixed point iteration
	\begin{equation}
		\theta^* \leftarrow \arg\min_\theta g(\theta,\theta^*)
	\end{equation}
	
	By definition, the iteration drives the objective function downhill. Strictly speaking, the descent property depends only on decreasing $g(\theta,\theta')$, not on strictly minimizing it. Under appropriate regularity conditions, an MM approach is guaranteed to converge to a stationary point of the objective function.
	\begin{equation}
		f(\theta'') \leq a \cdot g(\theta'',\theta') + b \leq a \cdot g(\theta',\theta') + b = f(\theta')
	\end{equation}

	\subsection{Iterated Probabilistic Control}
	
	Here we aim to point out that either of the optimal control objectives, {\color{black} $a$ and $m$, defined in (\ref{eq:SOC}) and (\ref{eq:RSOC}) respectively}, can be treated by means of the MM principle and by recycling the probabilistic control policies {\color{black} from section \ref{sec:probabilistic-control}}.
	
	First let us consider the SOC problem defined in (\ref{eq:SOC}). To set up an MM program we must find a function that majorizes the objective. Such is established by the following proposition.
	\begin{proposition} 
		\label{prop:MMSOC}
		Objective (\ref{eq:Iproj}) majorizes objective (\ref{eq:SOC}).
	\end{proposition}
	For the proof see \ref{sec:proof-of-proposition-refpropmmsoc}.
	
	By merit of the MM program the following fixed point iteration, if it exists, converges to the deterministic \textit{stochastic optimal control policy} defined as $\arg \min a$.
	\begin{equation}
		\label{eq:IFPI}
		\underline{\pi}_{T-1}^\bullet \leftarrow \arg\min_{\underline{\pi}_{T-1}} \kullbacks{p(\underline{\xi}_T;\underline{\pi}_{T-1})}{p^*(\underline{\xi}_T;\underline{\pi}^\bullet_{T-1})}
	\end{equation}
	
	Likewise we can consider the RSOC defined in (\ref{eq:RSOC}). A majorizing objective is established by the following proposition.
	\begin{proposition} 
		\label{prop:MMRSOC}
		Objective (\ref{eq:Mproj}) majorizes objective (\ref{eq:RSOC}). 
	\end{proposition}
	For the proof see \ref{sec:proof-of-proposition-refpropmmrsoc}.
	
	Again it follows that the following fixed point iteration, if it exists, converges to the deterministic \textit{risk sensitive optimal control policy} defined as $\arg \min b$.
	\begin{equation}
		\label{eq:MFPI}
		\underline{\pi}_{T-1}^\star \leftarrow \arg\min_{\underline{\pi}_{T-1}} 	\kullbacks{p^*(\underline{\xi}_T;\underline{\pi}^\star_{T-1})}{p(\underline{\xi}_T;\underline{\pi}_{T-1})}
	\end{equation}
	
\begin{figure}[t]
		\centering
		\includegraphics[width=.75\columnwidth]{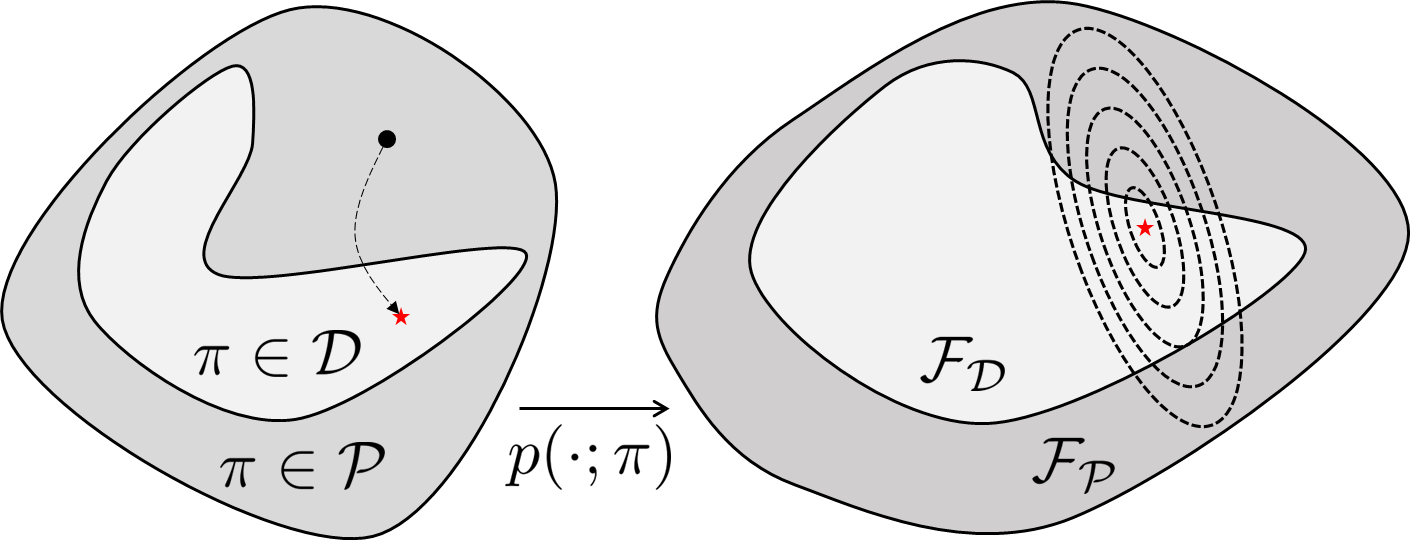}
		\caption{{\color{black}Illustration of the iterated probabilistic control policy. Represented on the left is the set of probabilistic policies, $\mathcal{P}$, with the subset of deterministic policies, $\mathcal{D}$. Represented on the right are the induced behaviour sets, $\mathcal{F}_\mathcal{D}$ and $\mathcal{F}_\mathcal{P}$. The level sets of the optimal control objective are represented with dotted lines. The optimal behaviour and corresponding policy are indicated with a red star. The iterated probabilistic control series is visualized by the dotted path.}}
		\label{fig:series}
	\end{figure}

	These propositions verify that the schemes in (\ref{eq:IFPI}) and (\ref{eq:MFPI}) determine two fixed point iterations. This procedure is visualized in Fig. \ref{fig:series}. {\color{black} The fixed point iterations suggest the existence of a series of policy sequences, say $\{\underline{\pi}_{T-1}^g\}$, with $g$ the element in the series, each occupying the set of probabilistic policies. The series converge to a} stationary point coinciding with the deterministic optimal control by merit of the MM principle. 
	{\color{black}The series of policy sequences can be initialized arbitrarily, as long as the support of the initial policy sequence contains the support of the deterministic optimal policy. Therewith the arbitrary policy sequence, $\underline{\rho}_T$, can be recognized as an intermediate result in the series. The closer the policy is to the optimal policy the fewer iterations are required to come arbitrarily close to it. These results are instrumental in the interpretation of probabilistic control as a generalization of optimal control. 
		
		We conclude this section with some final remarks.}
	
	{\color{black}\begin{remark}
			The associated fixed point iteration takes place in the extended optimization space spanned by the parametrized trajectory densities. The fact that we can establish an MM principle is a direct result of this extended optimization space. It is therefore tempting to interpret the functions, $\underline{\pi}_{T-1}^g$, as probability densities. From a technical point of view such an interpretation is irrelevant. Though here we argue that one may interpret each iterate policy sequence as a set of belief functions that express our diminishing epistemic uncertainty about the underlying deterministic solution of (\ref{eq:SOC}) either (\ref{eq:RSOC}).
		\end{remark}

		\begin{remark}
			It is interesting to note that the mode-covering M-projection is related to the risk-seeking RSOC problem where the closed-loop density will be more evenly spread. That whilst the mode-seeking I-projection is related to SOC.
	\end{remark}}

	{\color{black}\section{Moment projected probabilistic control revisited}\label{sec:bayesian-estimation--algorithms}}
	
	{\color{black} So far the treatment of either the information or moment projected probabilistic control formulation has been analogous. Though it is easily seen the recursion formula's are not equivalent and will induce different behaviour. In this section we point out that the moment projected probabilistic control problem holds a special position amongst either treatments. Therefore we demonstrate that} it is also possible to arrive at problems (\ref{eq:Mproj}) and (\ref{eq:MFPI}) using a different line of inquiry. {\color{black}It will proof worthwhile to include such an argument.} 
	
	\newpage
	
	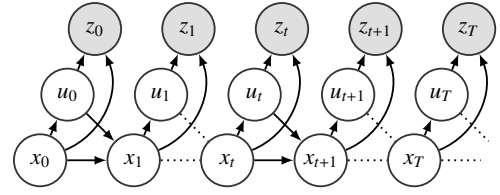
\begin{figure}[t] 
		\centering 
		\resizebox{.85\columnwidth}{!}{
			\begin{tikzpicture}
				\tikzstyle{main}=[circle, minimum size = 8mm, thick, draw =black!80, node distance = 6mm]
				\tikzstyle{connect} = [-latex,thick]
				\tikzstyle{floating} = [thick,dotted]
				\tikzstyle{measure} = [-latex,thick,bend right=45]
				
				\node[main] (X0) {$x_0$};
				\node[main] (X1) [right=of X0] {$x_1$};
				\node[main] (Xt) [right=of X1] {$x_t$};
				\node[main] (Xtt) [right=of Xt,label=center:$x_{t+1}$] {};
				\node[main] (XT) [right=of Xtt] {$x_{T}$};
				\node[main,draw=white] (XTT) [right=of XT] {};

				\node[main] (U0) [above left=of X1] {$u_0$};
				\node[main] (U1) [right=of U0] {$u_1$};
				\node[main] (Ut) [right=of U1] {$u_t$};
				\node[main] (Utt) [right=of Ut,label=center:$u_{t+1}$] {};
				\node[main] (UT) [right=of Utt] {$u_{T}$};
				
				\node[main,fill=gray!25] (Z0) [above left=of U1] {$z_0$};
				\node[main,fill=gray!25] (Z1) [right=of Z0] {$z_1$};
				\node[main,fill=gray!25] (Zt) [right=of Z1] {$z_t$};
				\node[main,fill=gray!25] (Ztt) [right=of Zt,label=center:$z_{t+1}$] {};
				\node[main,fill=gray!25] (ZT) [right=of Ztt] {$z_{T}$};
				
				\node[main,draw=white] (n1) [above left=of U0] {};
				
				\path (X0) edge [connect] (X1)
				(X1) edge [floating] (Xt)
				(Xt) edge [connect] (Xtt)
				(Xtt) edge [floating] (XT)
				(XT) edge [floating] (XTT);
				
				\path (X0) edge [measure] (Z0);
				\path (X1) edge [measure] (Z1);
				\path (Xt) edge [measure] (Zt);
				\path (Xtt) edge [measure] (Ztt);
				\path (XT) edge [measure] (ZT);
				
				\path (X0) edge [connect] (U0);
				\path (X1) edge [connect] (U1);
				\path (Xt) edge [connect] (Ut);
				\path (Xtt) edge [connect] (Utt);
				\path (XT) edge [connect] (UT);
				
				\path (U0) edge [connect] (X1);
				\path (U1) edge [floating] (Xt);
				\path (Ut) edge [connect] (Xtt);
				\path (Utt) edge [floating] (XT);
				\path (UT) edge [floating] (XTT);
				
				\path (U0) edge [connect] (Z0);
				\path (U1) edge [connect] (Z1);
				\path (Ut) edge [connect] (Zt);
				\path (Utt) edge [connect] (Ztt);
				\path (UT) edge [connect] (ZT);
			\end{tikzpicture}
		}
		\caption{Graphical representation of a Controlled Hidden Markov Model.} 
		\label{fig:CHMM} 
	\end{figure} 
	
	\subsection{Analogue Bayesian estimation problem}
	
	To that end we establish an analogy with Bayesian estimation \cite{sarkka2013bayesian}. This analogy will allows us to recycle the general treatment of a specific kind of Bayesian estimation problems {\color{black}and specialize them to our problem which in turn will proof equivalent to moment projected probabilistic control.} We extend the Controlled Markov Chain in Fig. \ref{fig:CMC}, to a Controlled Hidden Markov Model, introducing a measurement sequence, see Fig. \ref{fig:CHMM}. Assume that te measurements are independent of any history or future of the trajectory if the present state-action is known. If this is the case, the probability of a measurements, $\underline{z}_t = \{z_0,\dots,z_T\}$, conditioned on  trajectory, $\underline{\xi}_T$, is defined as
	\begin{equation} 
		p(\underline{z}_T|\underline{\xi}_t) = \prod\nolimits_{t=0}^{T} p(z_t|\xi_t) 
	\end{equation} 
	and the joint model parameterised by a policy sequence $\underline{\pi}_{T-1}$ as
	\begin{equation} 
		p(\underline{\xi}_t,\underline{z}_T;\underline{\pi}_{T-1}) = p(x_0)p(z_T|x_T)\prod\nolimits_{t=0}^{T-1} p(z_t|\xi_t)p(x_{t+1}|\xi_t)\pi_t(\xi_t) 
	\end{equation} 
	
	Clearly this model shares its main structure with that of a traditional Hidden Markov Model \cite{bishop2006pattern,pml1Book,sarkka2013bayesian}. Application of Bayesian inference procedures on this type of model are well studied and are easily extended to the present case. {\color{black} Two important examples are the Bayesian filtering problem and the Bayesian smoothing problem. With the latter we seek the marginal a posterior probability density, $p(\xi_t|\underline{z}_T;\underline{\pi}_{T-1})$.} 
	
	Then to establish a connection with probabilistic control the following emission model is proposed \cite{attias2003planning,toussaint2006probabilistic,toussaint2009robot,hoffmann2017linear}. We introduce an artificial binary measurement $z_t$ that we determine to have assumed the value $1$ with probability proportional to $e^{-c_t(\xi_t)}$. The associated local and joint measurement models are
	\begin{equation} 
		\begin{aligned} 
			&p(1|\xi_t) \propto e^{-c_t(\xi_t)} \\ 
			&p(\underline{1}_T|\underline{\xi}_T) \propto e^{-\underline{c}_T(\underline{\xi}_T)} 
		\end{aligned} 
	\end{equation}

	It is rather difficult to give a convincing justification for this model. Rather it should be understood as a technical trick. Because within this context the following Maximum Likelihood Estimation (MLE) problem is meaningful, assuming the graphical model in Fig. \ref{fig:CHMM} with \textit{optimal} observations $\underline{z}_T = \underline{1}_T$.
	\begin{equation} 
		\label{eq:MLE}
		\max_{\underline{\pi}_{T-1}} \log p(\underline{1}_T;\underline{\pi}_{T-1}) 
	\end{equation} 
	
	One easily verifies this problem is indeed equivalent to the RSOC problem defined in (\ref{eq:RSOC}) provided that
	\begin{equation} 
		\begin{aligned}
			p(\underline{1}_T;\underline{\pi}_{T-1}) &= \int p(\underline{1}_T|\underline{\xi}_T) p(\underline{\xi}_T;\underline{\pi}_{T-1}) \text{d}\underline{\xi}_T \\
			&\propto \int p(\underline{\xi}_T;\underline{\pi}_{T-1}) e^{-\underline{c}_T(\underline{\xi}_T)} \text{d}\underline{\xi}_T 
		\end{aligned}
	\end{equation} 
	
	This observation is remarkable in that it establishes a technical equivalence between the Bayesian argument developed here and the underlying RSOC problem. So we may treat this problem further by addressing it as a conventional Bayesian estimation problem rather than an optimal control problem. 
	
	{\color{black}
		\begin{remark}
			Note that this analogy is restricted to a risk-seeking setting where $\sigma>0$. This is because to establish the equivalence with the RSOC objective in (\ref{eq:RSOC}), the maximization in (\ref{eq:MLE}) has to correspond with the minimization of the negative objective in (\ref{eq:RSOC}). For $\sigma<0$ the equivalence deteriorates.   
	\end{remark}}
	
	{\color{black}\subsection{Bayesian inference on probabilistic graph models}}
	
	MLE problems of this form are usually not addressed directly, rather one treats them iteratively using the Expectation-Maximization (EM) algorithm. The EM algorithm is an application of the MM approach tailored to MLE problems. 
	
	Put briefly it can be shown that problem (\ref{eq:MLE}) is governed by the following fixed point iteration. 
	\begin{equation} 
		\label{eq:MLEMM} 
		\underline{\pi}_{T-1}^\star \leftarrow \arg \min_{\underline{\pi}_{T-1}\in\mathcal{P}}\kullbacks{p(\underline{\xi}_T|\underline{1}_T;\underline{\pi}_{T-1}^\star)}{p(\underline{\xi}_T;\underline{\pi}_{T-1})} 
	\end{equation} 
	{\color{black} We refer to \ref{sec:em-treatment-of-rsoc} for details.}
	
	The solution of the root problem is conveniently summarized by the following proposition. 		
	\begin{proposition} 
		\label{prop:EM}
		Consider the moment projection in (\ref{eq:MLEMM}) substituting $\underline{\rho}_{T-1}$ for $\underline{\pi}_{T-1}^\star$. Then the optimal control is given by
		\begin{equation*}
			\pi^\star_t(\xi_t) = \rho_t(\xi_t) \frac{p(\overline{1}_t|\xi_t;\overline{\rho}_{t+1})}{p(\overline{1}_t|x_t;\overline{\rho}_{t})} 
		\end{equation*}
		Defining
		\begin{equation*} 
			\begin{aligned} 
				Q_t^\star(\xi_t) &= -\log p(\overline{1}_{t}|\xi_t;\overline{\rho}_{t+1}) \\ 
				V^\star_t({x}_{t}) &= -\log p(\overline{1}_{t}|{x}_{t};\overline{\rho}_{t}) 
			\end{aligned} 
		\end{equation*} 
		and starting with $V^\star_T = \exp(-c_T(x_T))$, then the functions $V^\star_t$ and $Q_t^\star$ are generated recursively in a backward manner as follows
		\begin{equation*}
			\begin{aligned} 
				V^\star_{t}({x}_t) &= -\log \expect{\rho_t(u_t|x_t)}\left[\exp(-{Q}^\star_t(\xi_t))\right] \\
				Q^\star_{t}(\xi_t) &= c_t({\xi}_t)-\log \expect{p({x}_{t+1}|{\xi}_t)}\left[\exp(-V^\star_{t+1}({x}_{t+1}))\right] 
			\end{aligned} 
		\end{equation*} 
	\end{proposition}
	For the proof see \ref{sec:proof-op-proposition-refpropem}.
	
	One verifies these functions are governed by the exact same recursive expressions as described in proposition \ref{eq:Mproj}, illustrating both problems are indeed (technically) equivalent. {\color{black}Moreover, the functions, $V^\star_t$ and $Q_t^\star$, are recognized as backward messages on the probabilistic graph model \cite{pml2Book,pml1Book}.}
	
	Furthermore it shows that
	\begin{equation}
		\label{eq:BE}
		\pi_t^\star(\xi_t) \propto p(\xi_t|\underline{1}_T;\underline{\rho}_{T-1})
	\end{equation}
	implying that the control is directly related to the smoothing density.  This observation has interesting practical consequences given the well developed set of algorithms that are tailored to smoothing \cite{sarkka2013bayesian}. {\color{black}We further remark that, as far as we are aware of, this is the sole problem formulation associated to an optimal control formulation that truly can be determined by practising inference and more specifically by practising message passing on a probabilistic graph model.}
	
	{\color{black}\subsection{Path Integral Control solution}\label{sec:path-integral-control-solution}	We now bring to the attention a second interesting property of the solution of problem (\ref{eq:Mproj}) as specified in proposition \ref{prop:Mproj}. One verifies that
		\begin{equation}
			V_t^\star(x_t) = - \log \expect{p(\overline{\xi}_{t}|{x}_t;\overline{\rho}_{t})}\left[e^{- \underline{c}_T(\overline{\xi}_{t})}\right]
		\end{equation}
		
		In other words, the value function, $V_t^\star$, can be evaluated explicitly as the expectation of the exponentiated cost-to-go over the prior (future) trajectory density, $p(\underline{\xi}_T;\underline{\rho}_{T-1})$.	{\color{black}We emphasize this property holds only for the MM treatment of RSOC.} The MM scheme specified for SOC does not allow such a treatment. Though, since the SOC and RSOC problems are equivalent for deterministic dynamics, the properties extend generally to probabilistic treatment of deterministic optimal control problems. 

		There are several reasons why this an exciting observation. As far as we are aware of this is the sole problem associated to a generic optimal control problem that exhibits this particular behaviour. Analogous solutions have been known to exist for a restricted set of SOC problems known as Linearly Solvable Optimal Control (LSOC) \cite{kappen2012optimal,kappen2005linear,todorov2007linearly,dvijotham2012linearly,kappen2016adaptive}. In the context of LSOC such solutions have been referred to as Path Integral Control solutions. Second, this observation has interesting practical consequences. It shows that we can calculate $V_t^\star$ directly, e.g. Monte Carlo methods, circumventing the backward recursion.
		

	}

	{\color{black}\section{Exact solutions and algorithms}\label{sec:explicit-solutions}
		
		An appealing property featured by the MM iterations, (\ref{eq:SOC}) and (\ref{eq:RSOC}), to solve optimal control problems is that an explicit minimizer is accessible. Ergo, the nonlinear program, which causes great difficulty in practice, can be solved explicitly. 	On the other hand, it is still so that the equations in {\color{black}lemma} (\ref{prop:Iproj}) and propositions (\ref{prop:Mproj}) and (\ref{prop:Rproj}) can only be evaluated explicitly for a limited class of problems. {\color{black}Though these solutions may inspire treatment of more difficult problems and can serve as an inspiration for algorithmic development.}
		
		\subsection{Specialisation to linear-Gaussian models}\label{sec:specialization-to-linear-gaussian-models-and-quadratic-costs}
		
		For continuous space problems we can specialise the results to linear-Gaussian dynamic with Quadratic costs. Considering its generalising properties, we can concentrate on the Rény projection problem introduced in (\ref{eq:Rproj}). 
		
		\begin{proposition} 
			\label{prop:LQR}
			Define linear-Gaussian dynamics, $p(x_{t+1}|\xi_t) = \mathcal{N}(x_{t+1}|\matrixstyle{F}_{\xi,t} \xi_t + f_t,\matrixstyle{P}_t) $, and Quadratic costs, $c_t(\xi_t) = \tfrac{1}{2} \xi_t^\top \matrixstyle{C}_{\xi\xi,t} \xi_t +  \xi_t^\top \matrixstyle{C}_{\xi,t} + \dots$ and $c_T(x_T) =  \tfrac{1}{2} \xi_T^\top \matrixstyle{C}_{xx,T} x_T +  x_T^\top \matrixstyle{C}_{x,T} + \dots$ and linear-Gaussian prior policy densities, $\rho_t(\xi_t) = \mathcal{N}({u}_t|\matrixstyle{K}_t{x}_t +{k}_t,\Sigma_t)$. 
			
			Then the optimal control associated to (\ref{eq:Rproj}) is given by 
			\begin{equation*}
				\pi_t^*(\xi_t) = \mathcal{N}({u}_t|\matrixstyle{K}_t^*{x}_t +{k}_t^*,\Sigma_t^*)
			\end{equation*}
			
			Starting with $V^*_T = c_T$, the functions $V^*_t$ and $Q_T^*$ are than also quadratic
			\begin{equation*}
				\begin{aligned}
					V^*_t({x}_t) &= \tfrac{1}{2}\begin{bmatrix}
						1 \\ {x}_{t} 
					\end{bmatrix}^\top \begin{bmatrix}
						\square & V_{x,t}^{*,\top} \\
						V_{x,t}^{*} & V_{xx,t}^*
					\end{bmatrix}\begin{bmatrix}
						1 \\ {x}_{t} 
					\end{bmatrix}\\
					Q^*_t({\xi}_t) &= \tfrac{1}{2}\begin{bmatrix}
						1 \\ {\xi}_{t} 
					\end{bmatrix}^\top \begin{bmatrix}
						\square & Q_{\xi,t}^{*,\top}  \\
						Q_{\xi,t}^{*} & Q_{\xi\xi,t}^* \\
					\end{bmatrix}\begin{bmatrix}
						1 \\ {\xi}_{t}
					\end{bmatrix}\\
				\end{aligned}
			\end{equation*}
			where $\square$ indicates redundant (and unspecified) constants. The parameters are generated recursively in a backward manner
			\begin{equation*}
				\begin{aligned} 
					V_{x,t}^* &= Q_{x,t}^* + \matrixstyle{K}_t^{\top} \Sigma_t^{-1} {k}_t - \matrixstyle{K}_t^{*,\top} \Sigma_t^{*,-1} {k}_t^* \\
					V_{xx,t}^* &= Q_{xx,t}^* + \matrixstyle{K}_t^{\top} \Sigma_t^{-1} \matrixstyle{K}_t - \matrixstyle{K}_t^{*,\top} \Sigma_t^{*,-1} \matrixstyle{K}_t^* 
				\end{aligned}
			\end{equation*}
			and
			\begin{equation*}
				\begin{aligned} 
					Q^*_{\xi,t} &= C_{\xi,t} + \matrixstyle{F}_{\xi,t}^\top(V_{xx,t+1}^{*,-1}+\alpha\matrixstyle{P}_t)^{-1}\left(V_{xx,t+1}^{*,-1} V_{x,t+1}^*+{f}_t\right) \\
					Q_{\xi\xi,t}^* &= C_{\xi\xi,t} + \matrixstyle{F}_{\xi,t}^\top(V_{xx,t+1}^{*,-1}+\alpha\matrixstyle{P}_t)^{-1}\matrixstyle{F}_{\xi,t} 
				\end{aligned}
			\end{equation*}
			Finally the policy parameters are given by
			\begin{equation*}
				\begin{aligned}
					{k}_t^* &= \Sigma_t^*(\Sigma_t^{-1}{k}_t - Q^*_{u,t}) \\
					\matrixstyle{K}_t^* &= \Sigma_t^*(\Sigma_t^{-1}\matrixstyle{K}_t - Q^*_{ux,t}) \\
					\Sigma_t^* &= ( \Sigma_t^{-1} + Q^*_{uu,t})^{-1} 
				\end{aligned}
			\end{equation*}
		\end{proposition}
		For the proof see \ref{sec:proof-of-proposition-refproplqr}.
		
		This solution is analogous with the Linear Quadratic Gaussian (LQG) regulator and Linear Exponential Quadratic Gaussian (LEQG) regulator, respectively the SOC and RSOC version of the linear-quadratic problem established in the proposition.} {\color{black}Moreover, if the solution from proposition \ref{prop:LQR} is iterated, it converges to the LQG and LEQG solution.}
	
\begin{figure}[t]
		\centering
		\includegraphics[width=\columnwidth]{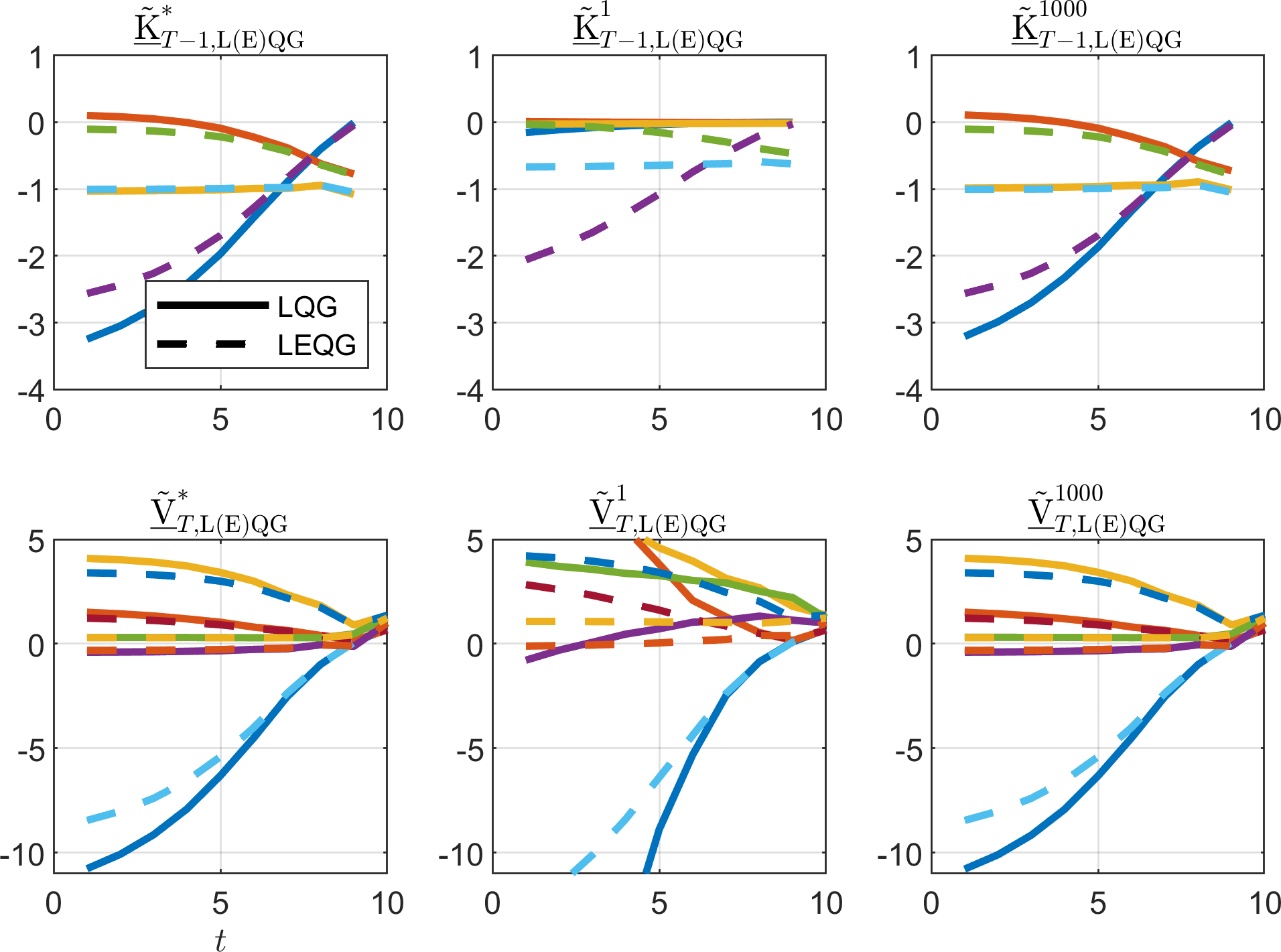}
		\caption{{\color{black}Comparison of the deterministic LQG and LEQR (left panels) and iterated probabilistic LQG and LEQG (middle and right panels) regulators. The middle and right panels show the probabilistic solution after one iteration and thousand iterations respectively. The top panels show the elements of the extended gain matrix, $\tilde{\matrixstyle{K}} = (k,\matrixstyle{K})$. The bottom panels show the upper triangular elements of the quadratic weight matrix of the value function.}}
		\label{fig:LQR}
	\end{figure}

	{\color{black}
		\begin{example}
			Consider as a simple example a random affine Gaussian system with problem dimensions $\dim(\mathcal{X}) = n_x = 2$ and $\dim(\mathcal{U}) = n_u = 1$ and horizon $T=10$ and with random quadratic cost. The solution of the L(E)QG regulator is then given by e.g. \cite{jacobson1973optimal}. The solution is compared with the iterated probabilistic controllers in figure \ref{fig:LQR}. The corresponding probabilistic controllers were initialised with the prior policy, $\rho_t = \mathcal{N}(0,10^{-2}), \forall t$. One verifies that the probabilistic control converges to the deterministic optimal control thereby illustrating the established connection between optimal and probabilistic control and the effectiveness of the fixed point iterations.
		\end{example}
		
	}

	\subsection{Towards algorithms}
	{\color{black} Based on the result from the previous section} we identify three interesting directions for future algorithmic development. We only focus on trajectory optimization. Other directions have been investigated in part by \Citealp{levine2018reinforcement,abdolmaleki2018maximum,haarnoja2017reinforcement,haarnoja2018soft} in the context of global solution strategies such as RL, albeit without assessing the possibility of establishing a fixed point iteration to {\color{black}assure that eventually the policy collapses onto the deterministic optimal controller}. {\color{black}As a core principle to each of the possible directions we have that the probabilistic control formulation, intrinsically, generates an iterative scheme whereas an iterative calculation is typically introduced ad hoc in other algorithms.}
	
	The first direction is general, the second and third direction relate to the MM treatment of RSOC problems {\color{black}and is thus only applicable to RSOC problems or extends to the SOC problem in the specific case that the dynamics are taken to be deterministic.}
	\begin{enumerate}
		\item The iterative Linear Quadratic Regulator (iLQR) \cite{todorov2005ilqr,mayne1966ddp,tassa2014control,howell2019ddp,theodorou2010stochastic} is an algorithm tailored to nonlinear trajectory optimization that makes use of the explicit LQR solution for linear quadratic (stochastic) optimal control problems. {\color{black}Specifically, in the forward pass, the policy is \textit{evaluated}, simultaneously linearizing the problem about the current nonlinear closed-loop trajectory. In a backward pass, an affine policy update is determined based on the LQR solution of the locally linearized problem.} Considering that the explicit expressions documented in section \ref{sec:specialization-to-linear-gaussian-models-and-quadratic-costs} are the probabilistic control analogues of the LQR solution, an analogues algorithm can be constructed tailored to SOC or RSOC problems that now consists of two iterations. In a main iteration the MM fixed point iterations is realised, in an inner iteration one could iterate a local linear approximation of the nonlinear probabilistic control problem in the exact same way as the iLQR algorithm does for nonlinear optimal control problems.
		\item We argue that with the treatment of RSOC problems through the MM principle and the associated property pointed out in section \ref{sec:path-integral-control-solution}, the concept of PIC is generalised to conventional optimal control problems other than the restricted class of LSOC problems. This approach establishes a theoretical basis to the wide variety of algorithms based on the concept and governing equations of PIC but that violate the condition that then $R$ can only depend on the state trajectory. This is most notably the case for the collection of Model Based PIC methods described in \cite{lefebvre2021deoc,stulp2012path,williams2016agressive,kahn2021land,kahn2021badgr,lefebvre2020elsoc,williams2018information,neve2022comparative}. Although these approaches are usually advanced as a stochastic solution approach tailored to SOC, the theory developed here demonstrates in fact the reciprocal RSOC is approximated. Furthermore this comparison complements the answer to the question posed by other authors who investigated the relation between LSOC and \textit{message passing} in the context of control as inference  \cite{watson2021control}.
		\item Finally it is also possible to exploit the connection with Bayesian estimation and the MM treatment of RSOC as was developed in section \ref{sec:bayesian-estimation--algorithms}. Equation (\ref{eq:BE}) demonstrates that the probabilistic policy, ${\underline{\pi}}_T^{\star}$, is determined by the input conditioned on the state and artificial measurements, $\underline{z}_T = \underline{1}_T$. So equivalently we can calculate the smoothing density, $p(\xi_t|\underline{1}_T;{\underline{\rho}}_T)$, using standard tools from Bayesian estimation and infer the probabilistic policy subsequently. The latter can then be used in the main fixed point iterations. For nonlinear problems these algorithms will iterate a local approximation of the exact smoothing density in proximity of some reference trajectory. There exist several numerical strategies to execute this procedure amongst which are the iterated Extended Rauch-Tung-Striebel (iERTS) and iterated Unscented Rauch-Tung-Striebel (iURTS). For details on smoothing we refer to \cite{sarkka2013bayesian}. Similar to the blue-print proposed in the first point, this idea would consist of a main and inner iteration. {\color{black}To emphasize the distinction with the algorithmic family proposed in the first point, we remark that iterated smoothers usually perform a forward filtering estimate first which is then corrected by a backward smoothing action. This would mean, contrary to iLQR and related approaches, optimality is encoded in the forward pass also instead of in the backward pass alone.}
	\end{enumerate}
	
	\vfill

	{\color{black}\section{Conclusion}\label{sec:discussion}
		
		In this contribution we presented a generalised view on probabilistic control where the goal is to find an uncertain or probabilistic control policy so that the closed-loop trajectory density is as close as possible to some desired reference density. {\color{black}We introduced an alternative parametrization of the reference density therewith exposing the close analogies with optimal control design formulations. We have also extended and discussed ways to measure the proximity of densities.} 
		
		The resulting probabilistic control problems can be solved explicitly yielding backward recursive expressions for the optimal probabilistic control policies. {\color{black}In contrast to optimal control design, the recursive expressions solely depend on the expectation operator suggestive of a particular ease when practising them.} Further, it was pointed out that the information or moment projected probabilistic control policies majorize Stochastic Optimal Control and Risk Sensitive Optimal Control problems respectively. This observation implies at two fixed point iterations that maintain a series of policy densities which can arguably be interpreted as our intermediate belief about the corresponding deterministic optimal controls. Furthermore, we highlighted that the moment projected probabilistic control approach apparently takes a special place amongst the other treatments. It is shown that the resulting solution allows for a Path Integral Control (PIC) expression circumventing the requirement to explicitly evaluate the entire recursive calculation. Furthermore, a technically equivalent treatment can be developed based on an analogy between a Bayesian MLE and the RSOC problem.
		
		\newpage
		
		{\color{black} The combination of these results establishes an explicit connection between probabilistic control and optimal control and provides theoretical grounding for any derived control architectures. The author believes that the value of these iterative schemes must be found in their intrinsically iterative exploration of the policy landscape. Although the optimal control policy is still to be preferred, the probabilistic control policy provides a tractable alternative that furthermore collapses onto the optimal policy when iterated. Reflecting on its use in Reinforcement Learning, the authors is not convinced about the probabilistic controllers capacity to incite useful or even goal-directed exploration in a context where also the system itself is unknown, a property that is often attributed to maximum entropy policies. However, it is noted that the moment projected probabilistic control formulation is the only control design approach where the value function, and thus indirectly the policy, can be evaluated through an expectation over closed-loop trajectories with a prior policy. This expectation can be evaluated using Monte Carlo methods, that is relying on (simulated) systems rollouts. This is a remarkable property that can be of value to many learning frameworks. 
			
			To conclude, the framework of probabilistic control undeniably exhibits potential but it remains to be established how to exploit its properties in practical algorithms.}
		
	}

		\section*{Acknowledgements}
		This work was supported by the Research Foundation Flanders (FWO) under SBO grant n° S007723N.
		
		\bibliographystyle{elsarticle-num}
		\bibliography{references}
		
		\vfill
		\appendix
		
		\section{Proof of lemma \ref{prop:Iproj}}\label{sec:proof-of-proposition-refpropiproj}
		
		\begin{proof}
			First verify that problem (\ref{eq:Iproj}) can be recast as
			\begin{equation}
				\min_{\underline{\pi}_{T-1} \in \mathcal{P}} \expect{p(\underline{\xi}_T;\underline{\pi}_{T-1})} \left[\underline{c}_T(\underline{\xi}_T)+\log\frac{\underline{\pi}_{T-1}(\underline{\xi}_T)}{\underline{\rho}_{T-1}(\underline{\xi}_T)}\right]
			\end{equation}
			
			Now remark that this problem exhibits an optimal substructure. Consider therefore
			\begin{equation}
				\begin{multlined}[.7\linewidth]
					\min_{\underline{\pi}_{T-1}\in\mathcal{P}} \int   \text{d}\underline{\xi}_T  p(\underline{\xi}_T;\underline{\pi}_{T-1})  \left(R(\underline{\xi}_T) + \frac{\underline{\pi}_{T-1}(\underline{\xi}_T)}{\underline{\rho}_{T-1}(\underline{\xi}_T)}\right) \\ =  \min_{\underline{\pi}_{t}\in\mathcal{P}} \int \text{d}\underline{\xi}_t \text{d}x_{t+1} p(\underline{\xi}_t,x_{t+1};\underline{\pi}_{t}) \left(\underline{c}_t(\underline{\xi}_t) + \log \frac{\underline{\pi}_{t}(\underline{\xi}_t)}{\underline{\rho}_{t}(\underline{\xi}_t)}\right)  V_{t+1}^\bullet(x_{t+1}) 
				\end{multlined}
			\end{equation}
			where 
			\begin{equation}
				V^\bullet_t(x_t) = \min_{\overline{\pi}_{t} \in \mathcal{P}} \expect{p(\overline{\xi}_t|x_t;\overline{\pi}_{t})} \left[\overline{c}_t(\overline{\xi}_t)+\log\frac{\overline{\pi}_{t}(\overline{\xi}_t)}{\overline{\rho}_{t}(\overline{\xi}_t)}\right]
			\end{equation}

			Further note that function, $V^\bullet_t$, satisfies a recursion, where $V_T^\bullet(x_T)$ is initialised as $\exp(-c_T(x_T))$.
			\newpage
			\begin{equation}
				\label{eq:Vbullet}
				\begin{multlined}[.85\linewidth]
					V^\bullet_t(x_t) \\ = \min_{\pi_t \in \mathcal{P}} \expect{p(u_t|x_t;\pi_t)} \left[ c_t(\xi_t) + \log \frac{\pi_t(\xi_t)}{\rho_t(\xi_t)} + \expect{p(x_{t+1}|\xi_t)}[V^\bullet_{t+1}(x_{t+1})]\right] 
				\end{multlined}
			\end{equation}
			
			The variational optimisation problem governing, $V^\bullet_t$, can be solved explicitly by making use of variational calculus. Consider the following Lagrangian. 
			
			The Lagrangian multiplier, $\lambda_t$, is associated to the normalization condition.
			\begin{equation}
				L_t = c_t(\xi_t) + \log \frac{\pi_t(\xi_t)}{\rho_t(\xi_t)} + \expect{p(x_{t+1}|\xi_t)}[V^\bullet_{t+1}(x_{t+1})] - \lambda_t \pi_t(\xi_t)
			\end{equation}
			
			It follows that
			\begin{equation}
				\begin{aligned}
					\log \pi_t(\xi_t) &= \log \rho_t(\xi_t) - c_t(\xi_t) - \expect{p(x_{t+1}|\xi_t)}[V^\bullet_{t+1}(x_{t+1})] + \lambda_t 
				\end{aligned}
			\end{equation}
			and thus
			\begin{equation}
				\pi_t(\xi_t) \propto \rho_t \exp(- c_t(\xi_t) - \expect{p(x_{t+1}|\xi_t)}[V^\bullet_{t+1}(x_{t+1})])
			\end{equation}
			
			Substitution of this expression into the definition of $V_t^\bullet$ and recognition of the function $Q_t^\bullet$ concluded the proof.	\qed
		\end{proof}
		
		\section{Proof of proposition \ref{prop:Mproj}}\label{sec:proof-of-proposition-refpropmproj}
		
		\begin{proof}
			Let us first further reduce problem (\ref{eq:Mproj}) 
			\begin{equation} 
				\label{eq:RSOC_iter1}
				\begin{multlined}[.85\linewidth]
					\arg\min_{\underline{\pi}_{T-1}\in\mathcal{P}}\kullbacks{p(\underline{\xi}_T;\underline{\rho}_{T-1})e^{-\underline{c}_T(\underline{\xi}_T)}}{p(\underline{\xi}_T;\underline{\pi}_{T-1})}  \\
					\begin{aligned} 
						&= \arg\min_{\underline{\pi}_{T-1}\in\mathcal{P}} \int p(\underline{\xi}_T;\underline{\rho}_{T-1})e^{-\underline{c}_T(\underline{\xi}_T)} \log \frac{p(\underline{\xi}_T;\underline{\rho}_{T-1})e^{-\underline{c}_T(\underline{\xi}_T)}}{p(\underline{\xi}_T;\underline{\pi}_{T-1})}\text{d}\underline{\xi}_T \\ 
						&= \arg\max_{\underline{\pi}_{T-1}\in\mathcal{P}} \int  p(\underline{\xi}_T;\underline{\rho}_{T-1})e^{\underline{c}_T(\underline{\xi}_T)}  \sum\nolimits_{t=0}^{T-1} \log \pi_t(\xi_t) \text{d}\underline{\xi}_T \\ 
						&= \arg\max_{\underline{\pi}_{T-1}\in\mathcal{P}}  \sum\nolimits_{t=0}^{T-1} \int  p(\underline{\xi}_T;\underline{\rho}_{T-1})e^{-\underline{c}_T(\underline{\xi}_T)} \log \pi_t(\xi_t) \text{d}\underline{\xi}_T 
					\end{aligned} 
				\end{multlined}
			\end{equation}
			
			This decomposition illustrates that as opposed to problem (\ref{eq:Iproj}), the present variational optimisation problem is not subject to an optimal substructure and can be treated independently for each individual policy. 
			
			Taking into account the normalisation condition it follows that 
			\begin{equation}
				\pi_t^\star(\xi_t)= \frac{v_t(\xi_t)}{\int v_t(\xi_t)\text{d}u_t}
			\end{equation}
			where
			\begin{equation}
				v_t(\xi_t) = \int \text{d}\underline{\xi}_{t-1}\text{d}\overline{\xi}_{t+1} p(\underline{\xi}_T;\underline{\rho}_{T-1})e^{-\underline{c}_T(\underline{\xi}_T)} 
			\end{equation}
			
			The function $v_t$ can be decomposed as follows
			\begin{equation}
				\begin{multlined}[.85\linewidth]
					v_t(\xi_t) \\
					\begin{aligned}
						&\begin{multlined}[.85\linewidth]
							\propto \rho_t(\xi_t)e^{-c_t(\xi_t)}\underbrace{\int \text{d}\underline{\xi}_{t-1}  p(\underline{\xi}_{t-1},x_t;\underline{\rho}_{t-1})e^{-\underline{c}_{t-1}(\underline{\xi}_{t-1})} }_{w_t(x_t)} \\
							\times \text{d}\overline{\xi}_{t+1} p(\overline{\xi}_{t+1}|\xi_t;\overline{\rho}_{t+1}) e^{-\overline{c}_{t+1}(\overline{\xi}_{t+1})}
						\end{multlined} \\
						&\propto \rho_t(\xi_t)e^{-c_t(\xi_t)}w_t(x_t)\int \text{d}\overline{\xi}_{t+1}  p(\overline{\xi}_{t+1}|\xi_t;\overline{\rho}_{t+1}) e^{-\overline{c}_{t+1}(\overline{\xi}_{t+1})}
					\end{aligned}
				\end{multlined}
			\end{equation}
			
			Now recall that the policy is established by normalisation on $u_t$, which implies that the factor, $w(x_t)$, occurs in both nominator and denominator and does not affect the result by consequence. We can then identify the recursive equations governing the solution.		\qed
		\end{proof}
		
		\section{Proof of proposition \ref{prop:Rproj}}\label{sec:proof-of-proposition-refproprproj}
		\begin{proof}
			Let us first further reduce problem (\ref{eq:Rproj}) 
			\begin{equation}
				\begin{multlined}
					\min_{\underline{\pi}_T} \int \text{d}\underline{\xi}_T p(\underline{\xi}_T;\underline{\rho}_T)^\alpha e^{-\alpha \underline{c}_T(\underline{\xi}_T)} p(\underline{\xi}_T;\underline{\pi}_T)^{1-\alpha}\\
					\begin{aligned}
						&
						= \min_{\underline{\pi}_T} \int \text{d}\underline{\xi}_T p(\underline{\xi}_T;\underline{\pi}_T^{1-\alpha})  e^{\overbrace{-\alpha \underline{c}_T(\underline{\xi}_T) + \alpha\log \underline{\rho}_T(\underline{\xi}_T)}^{-\alpha \underline{c}_T(\underline{\xi}_T;\underline{\rho}_T)}} \\
						&\begin{multlined}[.8\columnwidth]
							=  \min_{\underline{\pi}_t} \int \text{d}\underline{\xi}_t \text{d}x_{t+1} p(\underline{\xi}_t,x_{t+1};\underline{\pi}_t^{1-\alpha}) e^{-\alpha \underline{c}_t(\underline{\xi}_t;\underline{\rho}_t)} \\
							\times \min_{\overline{\pi}_{t+1}} \int \text{d}u_{t+1}\text{d}\overline{\xi}_{t+2}  p(\overline{\xi}_{t+1}|x_{t+1};\overline{\pi}_{t+1}^{1-\alpha}) e^{-\alpha \overline{c}_{t+1}(\overline{\xi}_{t+1};\overline{\rho}_{t+1})}
						\end{multlined} \\
						&= \min_{\underline{\pi}_t} \int \text{d}\underline{\xi}_t  p(\underline{\xi}_t,x_{t+1};\underline{\pi}_t^{1-\alpha}) e^{-\alpha \underline{c}_t(\underline{\xi}_t;\underline{\rho}_t)} Z_{t+1}(x_{t+1})
					\end{aligned}
				\end{multlined}
			\end{equation}
			where 
			\begin{equation}
				\begin{multlined}
					Z_{t}(x_{t}) \\
					\begin{aligned}
						&= \min_{\overline{\pi}_{t}} \int \text{d}u_{t}\text{d}\overline{\xi}_{t+1}  p(\overline{\xi}_{t}|x_{t};\overline{\pi}_{t}^{1-\alpha}) e^{-\alpha \underline{c}_t(\overline{\xi}_{t};\overline{\rho}_{t})} \\
						&\begin{multlined}[.8\columnwidth]
							=\min_{{\pi}_{t}} \int \text{d}u_{t}\text{d}{x}_{t+1} p(x_{t+1}|\xi_t)  \pi_t(\xi_t)^{1-\alpha} e^{-\alpha c_t(\xi_t;\rho_t)} \\ \times \min_{\overline{\pi}_{t+1}} \int \text{d}u_{t+1}\text{d}\overline{\xi}_{t+2}  p(\overline{\xi}_{t+1}|x_{t+1};\overline{\pi}_{t+1}^{1-\alpha}) e^{-\alpha \overline{c}_{t+1}(\overline{\xi}_{t+1};\overline{\rho}_{t+1})}
						\end{multlined} \\
						&\begin{multlined}[.8\columnwidth]
							=\min_{{\pi}_{t}} \int \text{d}u_{t}\text{d}{x}_{t+1} p(x_{t+1}|\xi_t) \pi_t(\xi_t)^{1-\alpha} e^{-\alpha c_t(\xi_t;\rho_t)} \\ \times Z_{t+1}(x_{t+1})
						\end{multlined}
						\\
					\end{aligned}
				\end{multlined}
				\vspace*{-10pt}
			\end{equation}
			
			The latter can be solved for $\pi_t$ using variational calculus and taking into account the normalization, which yields
			\begin{equation}
				\pi_t \propto \rho_t e^{-c_t} \expect{}[Z_{t+1}]^{\frac{1}{\alpha}}
			\end{equation}
			
			Let us now define $Z_t = \exp(-\alpha V_t)$ and
			\begin{equation}
				Q_t = c_t - \frac{1}{\alpha }\log   \expect{}[\exp(-\alpha V_{t+1})]
			\end{equation}
			then one verifies that
			\begin{equation}
				\begin{aligned}
					\pi_t &= \rho_t \frac{\exp(-Q_t)}{\exp(-V_t)} \\
					V_t &= - \log \expect{\rho_t}[\exp(-Q_t)]
				\end{aligned}
			\end{equation} 
			concluding the proof. \qed
		\end{proof}

		\section{Proof of proposition \ref{prop:MMSOC}}\label{sec:proof-of-proposition-refpropmmsoc}
		\begin{proof}
			Consider the following analysis
			\begin{equation}
				\label{eq:decomp1}
				\begin{multlined}
					a[\underline{\pi}_{T-1}] \\
					\begin{aligned}
						&= \int p(\underline{\xi}_T;\underline{\pi}_{T-1}) \underline{c}_T(\underline{\xi}_T)\text{d}\underline{\xi}_T \\
						& = \int p(\underline{\xi}_T;\underline{\pi}_{T-1}) \log \frac{1}{e^{-\underline{c}_T(\underline{\xi}_T)}} \text{d}\underline{\xi}_T \\
						&\begin{multlined}[.9\columnwidth]
							= \int p(\underline{\xi}_T;\underline{\pi}_{T-1}) \log \frac{p(\underline{\xi}_T;\underline{\pi}_{T-1})}{p(\underline{\xi}_T;\underline{\rho}_{T-1})e^{-\underline{c}_T(\underline{\xi}_T)}} \text{d}\underline{\xi}_T \\+ \int p(\underline{\xi}_T;\underline{\pi}_{T-1}) \log \frac{p(\underline{\xi}_T;\underline{\rho}_{T-1}) }{p(\underline{\xi}_T;\underline{\pi}_{T-1})} \text{d}\underline{\xi}_T 
						\end{multlined}\\
						&\begin{multlined}[.9\columnwidth]= \kullbacks{p(\underline{\xi}_T;\underline{\pi}_{T-1})}{p(\underline{\xi}_T;\underline{\rho}_{T-1})e^{-\underline{c}_T(\underline{\xi}_T)}} \\-\underbrace{\kullbacks{p(\underline{\xi}_T;\underline{\pi}_{T-1})}{p(\underline{\xi}_T;\underline{\rho}_{T-1})}}_{\geq 0}\end{multlined}\\
						&\leq  a\cdot \kullbacks{p(\underline{\xi}_T;\underline{\pi}_{T-1})}{p^*(\underline{\xi}_T;\underline{\rho}_{T-1})} + b
					\end{aligned}
				\end{multlined}
			\end{equation}
			which illustrates that (\ref{eq:Iproj}) in fact majorizes $a[\underline{\pi}_{T-1}]$. \qed
		\end{proof}
		
		\section{Proof of proposition \ref{prop:MMRSOC}}\label{sec:proof-of-proposition-refpropmmrsoc}
		\begin{proof}
			Consider the following analysis
			\begin{equation}
				\begin{multlined}
					m[\underline{\pi}_{T-1}] \\
					\begin{aligned}
						&= -\log \int p(\underline{\xi}_T;\underline{\pi}_{T-1}) e^{-\underline{c}_T(\underline{\xi}_T)} \text{d}\underline{\xi}_T \\
						&= -\log \eta[\underline{\pi}_{T-1}] \\
						&=  -\int \tfrac{1}{\eta[\underline{\rho}_{T-1}]}  p(\underline{\xi}_T;\underline{\rho}_{T-1})e^{-\underline{c}_T(\underline{\xi}_T)}\log\frac{1}{\tfrac{1}{\eta[\underline{\pi}_{T-1}]}}\text{d}\underline{\xi}_T\\
						&\begin{multlined}[.85\linewidth]
							=  \int \tfrac{1}{ \eta[\underline{\rho}_{T-1}]}p(\underline{\xi}_T;\underline{\rho}_{T-1}) e^{-\underline{c}_T(\underline{\xi}_T)} \\
							\times \log \frac{\frac{1}{\eta[\underline{\rho}_{T-1}]}p(\underline{\xi}_T;\underline{\rho}_{T-1}) e^{-\underline{c}_T(\underline{\xi}_T)}}{ p(\underline{\xi}_T;\underline{\pi}_{T-1}) e^{-\underline{c}_T(\underline{\xi}_T)}}\text{d}\underline{\xi}_T \\ - \int \tfrac{1}{ \eta[\underline{\rho}_{T-1}]}  p(\underline{\xi}_T;\underline{\rho}_{T-1}) e^{-\underline{c}_T(\underline{\xi}_T)} \\
							\times \log \frac{ \frac{1}{\eta[\underline{\rho}_{T-1}]}p(\underline{\xi}_T;\underline{\rho}_{T-1}) e^{-\underline{c}_T(\underline{\xi}_T)}}{\frac{1}{\eta[\underline{\pi}_{T-1}]}p(\underline{\xi}_T;\underline{\pi}_{T-1}) e^{-\underline{c}_T(\underline{\xi}_T)}}\text{d}\underline{\xi}_T
						\end{multlined} \\
						&\begin{multlined}[.85\linewidth]
							= \kullback\left[\tfrac{1}{\eta[\underline{\rho}_{T-1}]}p(\underline{\xi}_T;\underline{\rho}_{T-1}) e^{-\underline{c}_T(\underline{\xi}_T)}\left\|p(\underline{\xi}_T;\underline{\pi}_{T-1}) e^{-\underline{c}_T(\underline{\xi}_T)}\right]\right. \\ - \underbrace{\kullback\left[\tfrac{1}{\eta[\underline{\rho}_{T-1}]}p(\underline{\xi}_T;\underline{\rho}_{T-1}) e^{-\underline{c}_T(\underline{\xi}_T)}\left\|\tfrac{1}{\eta[\underline{\pi}_{T-1}]}p(\underline{\xi}_T;\underline{\pi}_{T-1}) e^{-\underline{c}_T(\underline{\xi}_T)}\right]\right.}_{\geq 0}
						\end{multlined} \\
						&\leq a \cdot \kullback\left[p^*(\underline{\xi}_T;\underline{\rho}_{T-1}) \left\|p(\underline{\xi}_T;\underline{\pi}_{T-1}) \right]\right. + b
					\end{aligned}
				\end{multlined}
			\end{equation}
			which illustrates that (\ref{eq:Mproj}) in fact majorizes $m[\underline{\pi}_{T-1}]$. \qed
		\end{proof}

		\section{EM treatment of RSOC}\label{sec:em-treatment-of-rsoc}
		
		Consider a probabilistic model with hidden and observable variables, $X$ and $Z$ (for an example we refer to Fig. \ref{fig:CHMM}). Further suppose that the probabilistic model, $\mathcal{M}$, is characterised by a set of variables $\theta$, so that the joint density is given by, $p(Z,X;\theta)$. The Maximum Likelihood Estimation (MLE) objective is to identify parameters, $\theta$, by maximises the likelihood of observations, $Z$
		\begin{equation}
			\max_\theta \log \int p(Z|X;\theta)p(X;\theta)\text{d}X
		\end{equation}
		
		It is well-known that this objective is difficult to treat on account of the integral expression. To circumvent the intractable inference, an auxiliary {inference density}, $q(X)$, is introduced. The inference density allows to decompose the objective into a surrogate objective, $\mathcal{L}$, the so called evidence lower bound (ELBO), and, a relative entropy \textit{error} term. Since the relative entropy is positive semi-definite, with equality only if $q(X) \equiv p(X|Z;\theta)$, the ELBO minorizes the $\log$ likelihood and can be used to establish an MM program.
		
		\begin{equation}
			\begin{multlined}
				\log p(Z;\theta) \\
				\begin{aligned}
					&= \int q(X) \log \frac{p(Z,X;\theta)}{q(X)}\text{d}X + \int q(X) \log \frac{q(X)}{p(X|Z;\theta)} \text{d}X \\
					&= \mathcal{L}[q(X)|Z;\theta] + \underbrace{\mathbb{D}\left[q(X)||p(X|Z;\theta)\right]}_{\geq 0} \\
					&\geq \mathcal{L}[q(X)|Z;\theta]
				\end{aligned}
			\end{multlined}
		\end{equation}
		
		In theory the inference density can be chosen arbitrarily. As long as the domination and tangency condition are satisfied (\ref{eq:MM}), the ELBO can be considered as a minorizing surrogate. Though the sole inference density that satisfies these two conditions is the a posteriori density $p(X|Z;\theta)$. The EM algorithm is summarized by the following fixed point iteration
		\begin{equation}
			\theta^* \leftarrow \expect{p(X|Z;\theta^*)}[\log p(X,Z;\theta)]
		\end{equation}
		
		Finally note that
		\begin{equation}
			\expect{p(X|Z;\theta^*)}[\log p(X,Z;\theta)] = a\cdot \kullbacks{p(X|Z;\theta^*)}{p(X,Z;\theta)} + b
		\end{equation}
		
		\section{Proof op proposition \ref{prop:EM}}\label{sec:proof-op-proposition-refpropem}
		\begin{proof}
			The treatment of problem (\ref{eq:MLEMM} ) is analogues to the first part of the proof of proposition \ref{prop:Mproj}, (see \ref{sec:proof-of-proposition-refpropmproj}. It follows 
			\begin{equation} 
				\pi^\star_t(\xi_t) = \frac{p(\xi_t|\underline{1}_T;\underline{\rho}_{T-1})}{p(x_t|\underline{1}_T;\underline{\rho}_{T-1})} = p(u_t|x_t,\underline{1}_T;\underline{\rho}_{T-1}) 
			\end{equation} 
			The solution is thus equivalent to the probability of $u_t$ conditioned onto the measurements, $\underline{z}_T$, as parametrized by the prior policy sequence, $\underline{\rho}_{T-1}$.
			
			We can submit the densities, $\underline{\pi}^\star_T$, to further assessment.  Since we conditioned on the state ${x}_t$, it follows that this density is equivalent to $p({u}_t|{x}_t,\overline{1}_t;\overline{\rho}_{t-1})$ rather than $p({u}_t|{x}_t,\underline{1}_T;\underline{\rho}_{T-1})$. This is a direct result of the Markov property, since no more information about the input at time instant $t$ can be contained in the historical (and fictitious) measurements, $\underline{z}_{t-1} = \underline{1}_{t-1}$, than is already contained by the state, ${x}_t$, itself. This resonates with the common sense that once we have arrived at some state, ${x}_t$, we can only hope to reproduce the measurements, $\overline{z}_t = \overline{1}_t$, but can no longer affect any of the preceding measurements, $\underline{z}_{t-1} = \underline{1}_{t-1}$. 
			
			Now using Bayes' rule, $\pi^\star_t$ can be decomposed as 
			\begin{equation} 
				p({u}_t|{x}_t,\overline{1}_t;\overline{\rho}_{t}) = \rho_t(\xi_t) \frac{p(\overline{1}_t|\xi_t;\overline{\rho}_{t+1})}{p(\overline{1}_t|x_t;\overline{\rho}_{t})} 
			\end{equation} 
			which reduces the problem to finding efficient expressions for the probabilities $p(\overline{1}_t|x_t;\overline{\rho}_{t})$ and $p(\overline{1}_t|\xi_t;\overline{\rho}_{t+1})$. The latter is a generalisation of the backward filtering densities.
			
			We have that
			\begin{equation} 
				p(\overline{1}_t|x_t;\overline{\rho}_{t}) = \int {\rho}_{t}(\xi) p(\overline{1}_t|\xi_t;\overline{\rho}_{t+1})\text{d}u_t = \expect{{\rho}_{t}(\xi_t)}\left[p(\overline{1}_t|\xi_t;\overline{\rho}_{t+1})\right] 
			\end{equation} 
			These backward filtering densities adhere to a backward recursive expression. 
			\begin{equation} 
				\begin{aligned} 
					p(\overline{1}_t|\xi_t;\overline{\rho}_{t+1}) &= p(1|\xi_t) \int p(x_{t+1}|\xi_t) p(\overline{1}_{t+1}|x_{t+1};\overline{\rho}_{t+1})\text{d}x_{t+1} \\ 
					&= p(1|\xi_t) \expect{p(x_{t+1}|\xi_t)}\left[p(\overline{1}_{t+1}|x_{t+1};\overline{\rho}_{t+1})\right] 
				\end{aligned} 
			\end{equation} 
			
			Now define the functions  
			\begin{equation} 
				\begin{aligned} 
					Q_t^\star(\xi_t) &= -\log p(\overline{1}_{t}|\xi_t;\overline{\rho}_{t+1}) \\ 
					V^\star_t({x}_{t}) &= -\log p(\overline{1}_{t}|{x}_{t};\overline{\rho}_{t}) 
				\end{aligned} 
			\end{equation}
			
			One then easily verifies these functions satisfy the required recursive expressions.
			\qed
		\end{proof}
		
		\section{Proof of proposition \ref{prop:LQR}}\label{sec:proof-of-proposition-refproplqr}
		\begin{proof}
			In the present context it reasonable to hypothesize that both $Q_t^*$ and $V_t^*$ will be quadratic in their arguments
			\begin{equation}
				\begin{aligned}
					V^*_t({x}_t) &= \tfrac{1}{2}\begin{bmatrix}
						1 \\ {x}_{t} 
					\end{bmatrix}^\top \begin{bmatrix}
						\square & V_{x,t}^{*,\top} \\
						V_{x,t}^{*} & V_{xx,t}^*
					\end{bmatrix}\begin{bmatrix}
						1 \\ {x}_{t} 
					\end{bmatrix}\\
					Q^*_t({\xi}_t) &= \tfrac{1}{2}\begin{bmatrix}
						1 \\ {\xi}_{t} 
					\end{bmatrix}^\top \begin{bmatrix}
						\square & Q_{\xi,t}^{*,\top}  \\
						Q_{\xi,t}^{*} & Q_{\xi\xi,t}^* \\
					\end{bmatrix}\begin{bmatrix}
						1 \\ {\xi}_{t}
					\end{bmatrix}\\
				\end{aligned}
			\end{equation}
			
			Since $\pi_t^*(\xi_t) \propto \rho_t(\xi_t) \exp(-Q^*_t(\xi_t))$ one verifies that
			\begin{equation}
				\begin{aligned}
					{k}_t^* &= \Sigma_t^*(\Sigma_t^{-1}{k}_t - Q^*_{u,t}) \\
					\matrixstyle{K}_t^* &= \Sigma_t^*(\Sigma_t^{-1}\matrixstyle{K}_t - Q^*_{ux,t}) \\
					\Sigma_t^* &= ( \Sigma_t^{-1} + Q^*_{uu,t})^{-1} 
				\end{aligned}
			\end{equation}
			
			Further, as $\exp(-V^*_t(x_t)) \pi_t^*(\xi_t) = \rho_t(\xi_t) \exp(-Q^*_t(\xi_t))$ for any $u_t$ including $0$, it follows that
			\begin{equation}
				\label{eq:V-ELQR}
				\begin{aligned} 
					V_{x,t}^* &= Q_{x,t}^* + \matrixstyle{K}_t^{\top} \Sigma_t^{-1} {k}_t - \matrixstyle{K}_t^{*,\top} \Sigma_t^{*,-1} {k}_t^* \\
					V_{xx,t}^* &= Q_{xx,t}^* + \matrixstyle{K}_t^{\top} \Sigma_t^{-1} \matrixstyle{K}_t - \matrixstyle{K}_t^{*,\top} \Sigma_t^{*,-1} \matrixstyle{K}_t^* 
				\end{aligned}
			\end{equation}
			
			Finally one can verify the expression for $Q_t^*$ through substitution. \qed
		\end{proof}

	\end{document}